\documentclass[twoside]{article}

\usepackage{PRIMEarxiv}

\usepackage[ruled,vlined]{algorithm2e} 
\usepackage{subcaption}
\usepackage{amsmath}
\usepackage{caption}
\usepackage[utf8]{inputenc} 
\usepackage[T1]{fontenc}    
\usepackage{hyperref}       
\usepackage{url}            
\usepackage{booktabs}       
\usepackage{amsfonts}       
\usepackage{microtype}      
\usepackage{fancyhdr}       
\usepackage{graphicx}       
\graphicspath{{media/}}     

\newcommand{\fulltitle}{Impact of Price Inflation on Algorithmic Collusion Through Reinforcement Learning Agents}

\title{\fulltitle}

\author{
  Sebastián Tinoco \\
  Center for Artificial Intelligence (CENIA)\\
  University of Chile \\
  Santiago, Chile\\
  \texttt{stinoco@fen.uchile.cl} \\
   \And
  Andrés Abeliuk \\
  Department of Computer Sciences and\\
   Center for Artificial Intelligence (CENIA)\\
  University of Chile \\
  Santiago, Chile\\
  \texttt{aabeliuk@dcc.uchile.cl} \\
   \And
  Javier Ruiz del Solar\\
  Department of Electrical Engineering and\\
  Advanced Mining Technology Center (AMTC) \\
  University of Chile \\
  Santiago, Chile\\
  \texttt{jruizd@ing.uchile.cl} \\
}

\begin{document}
\maketitle

\begin{abstract}
Algorithmic pricing is increasingly shaping market competition, raising concerns about its potential to compromise competitive dynamics. While prior work has shown that reinforcement learning (RL)-based pricing algorithms can lead to tacit collusion, less attention has been given to the role of macroeconomic factors in shaping these dynamics. This study examines the role of inflation in influencing algorithmic collusion within competitive markets. By incorporating inflation shocks into a RL-based pricing model, we analyze whether agents adapt their strategies to sustain supra-competitive profits. Our findings indicate that inflation reduces market competitiveness by fostering implicit coordination among agents, even without direct collusion. However, despite achieving sustained higher profitability, agents fail to develop robust punishment mechanisms to deter deviations from equilibrium strategies. The results suggest that inflation amplifies non-competitive dynamics in algorithmic pricing, emphasizing the need for regulatory oversight in markets where AI-driven pricing is prevalent.
\end{abstract}


\section{Introduction}
\begin{figure}[h]
  \centering
  \includegraphics[width=12cm]{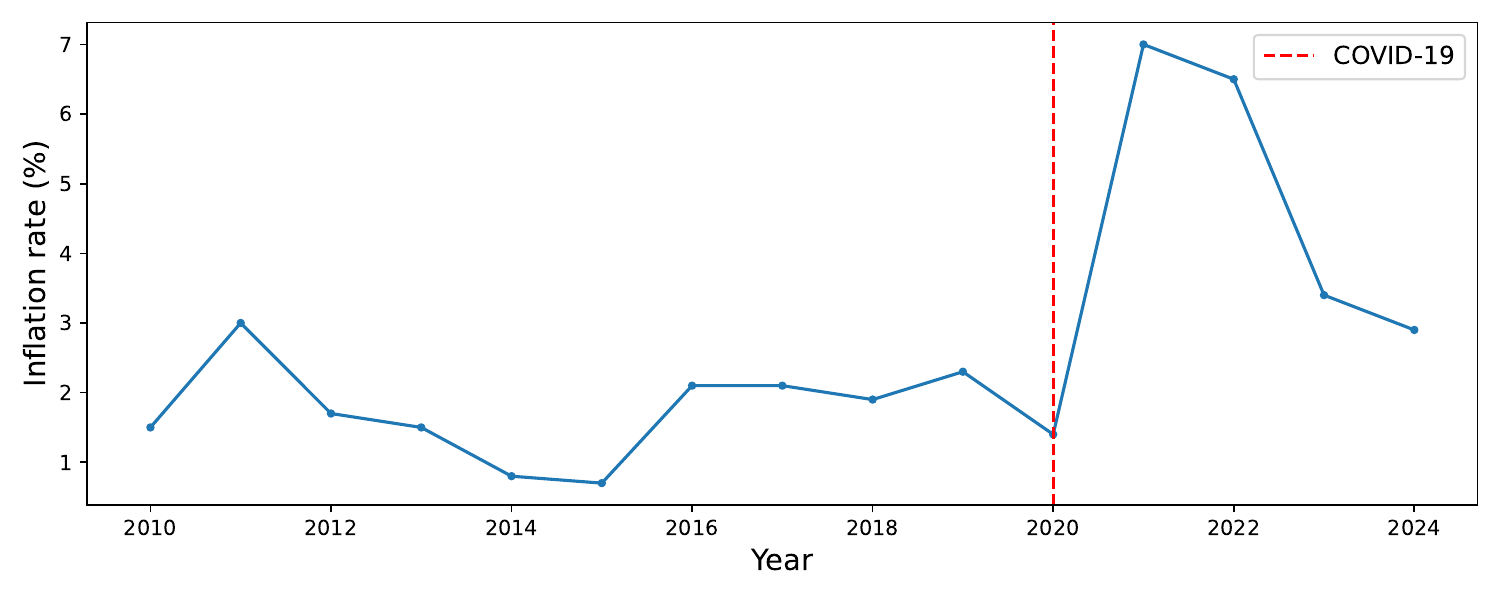}
  \caption{Annual Inflation Rate in the United States, 2010--2024.}
  \label{fig:inflation}
\end{figure}

Inflation has reemerged as a central concern in economic policymaking and business strategy, with developed economies experiencing sustained increases in price levels over the past decade. This shift, partly driven by expansive monetary responses to the COVID-19 pandemic, has heightened interest in how inflation affects firm behavior, particularly with respect to pricing. As illustrated in Figure~\ref{fig:inflation}, the annual inflation rate in the United States has markedly exceeded its long-term average in recent years, introducing greater volatility into markets traditionally assumed to be stable.

Inflation introduces both cost-side and demand-side uncertainties that influence firms’ pricing decisions. Prior work has shown that inflation can lead to greater price dispersion~\cite{chirinko2000market}, reduce price informativeness, and amplify firms’ pricing power—particularly in environments with high search or switching costs~\cite{janger2010relationship, gwin2004role}. These dynamics are well-recognized in classical economic theory~\cite{sheshinski1977inflation}, yet their interaction with modern algorithmic pricing systems remains underexplored.

At the same time, a growing body of literature has examined how reinforcement learning (RL) algorithms deployed in competitive markets can learn to coordinate on supra-competitive pricing without explicit communication. These findings have raised important questions about the susceptibility of algorithmic systems to tacit collusion, particularly in dynamic market environments. However, most existing models abstract away from macroeconomic fluctuations, assuming stationary cost and demand conditions that may not reflect the environments in which algorithmic pricing is increasingly deployed.

This paper addresses this gap by examining how inflationary shocks affect the learning dynamics and strategic behavior of autonomous pricing agents. By incorporating macroeconomic volatility into a standard RL-based pricing framework, we seek to better understand how external economic forces shape the emergence and stability of algorithmic coordination in competitive markets. In doing so, we aim to contribute to a more comprehensive framework for analyzing algorithmic interactions under realistic economic conditions.

\subsection*{Contributions}

This paper contributes to the literature on algorithmic pricing and macroeconomic dynamics by introducing and analyzing the role of inflation in environments governed by autonomous pricing agents. The key contributions are as follows:

\begin{itemize}
    \item We extend existing reinforcement learning pricing models by incorporating exogenous inflation shocks, enabling the study of algorithmic behavior under macroeconomic volatility.
    
    \item We propose a novel performance metric that accounts for inflation-adjusted profitability, allowing for a more realistic evaluation of strategic pricing outcomes in dynamic economic settings.
    
    \item We provide empirical simulation evidence on the behavior of deep reinforcement learning agents in inflationary contexts, contributing to the understanding of how non-stationary environments shape competitive dynamics.
\end{itemize}

\section{Related work}

The possibility of tacit \emph{algorithmic collusion}—where autonomous pricing algorithms learn to sustain supra-competitive prices without explicit communication—has attracted significant interest across economics and computer science. A foundational study by Calvano et al.~\cite{calvano} provides initial evidence that Q-learning~\cite{qlearning} agents competing in a repeated differentiated Bertrand~\cite{bertrand} duopoly can converge to supra-competitive pricing strategies, enforced through “punishment” phases after deviations. This finding has sparked a growing literature on the emergence of collusion in algorithmic markets.

Subsequent work has expanded these results across various market conditions. Calvano et al.~\cite{calvano2021} demonstrate that even under imperfect monitoring—akin to the Green–Porter framework~\cite{green}—Q-learning agents sustain collusion via finite trigger strategies. Klein~\cite{klein} shows that even simple learners can coordinate through Edgeworth price cycles, while Banchio and Mantegazza~\cite{banchio2022} propose a theoretical model in which spontaneous synchronization of experimentation schedules leads to implicit coordination. Hansen et al.~\cite{hansen2021} and Brown and MacKay~\cite{brown2023} further show that correlated pricing behavior may arise when agents ignore or adapt to rivals' actions, often without intentional coordination. Eschenbaum~\cite{eschenbaum} extends these findings by exploring the robustness of learned collusion under varying noise and uncertainty levels.

On the empirical side, Assad et al.~\cite{assad2024} study the German retail gasoline market and find that algorithmic pricing adoption leads to higher margins when both firms in a local duopoly use pricing software. Musolff~\cite{musolff2022} reports similar patterns in e-commerce, where algorithmic repricing reduces price dispersion and aligns sellers’ behaviors.

Parallel strands investigate interventions to limit algorithmic collusion. Johnson, Rhodes, and Wildenbeest~\cite{johnson2023} model how platform design—such as demand steering or ranking rules—can restore competitive pricing. Brero et al.~\cite{brero2022}, using a Stackelberg reinforcement learning framework~\cite{stackelberg1934}, show that a meta-algorithm can shape agent incentives to prevent coordination by penalizing high prices. These works suggest that market design and platform policies can be effective tools to mitigate undesired emergent behavior.

Recent contributions have also begun exploring the role of large language models (LLMs). Fish et al.~\cite{fish2024algorithmic} fine-tune LLMs on pricing tasks and observe that these models can autonomously generate natural-language policies that sustain collusive outcomes, highlighting new concerns about transparency and regulation in AI-driven markets.

In contrast to prior work, which predominantly models algorithmic pricing in stationary environments, our study incorporates exogenous macroeconomic shocks—specifically inflation—into the learning environment. This departs from existing approaches that assume stable cost and demand conditions, allowing us to investigate how external economic volatility interacts with agents' adaptive behavior. To our knowledge, this is the first study to examine the intersection of macroeconomic instability and reinforcement learning in pricing.

\section{Proposed method}

Section \ref{adding_inflation} introduces our approach to incorporating inflation into price competition dynamics. In Section \ref{mdp}, we formalize the Markov Decision Process (MDP) that governs agent decision-making in the pricing environment. Finally, Section \ref{section:agent} provides a detailed explanation of the reinforcement learning algorithm used by the agents to adapt their pricing strategies.

\subsection{Adding Inflation to Price Competition} \label{adding_inflation}

We model market dynamics through a Bertrand competition framework with differentiated products \cite{singh1984price}, employing its logit formulation. The quantity \( q_{i,t} \) sold by agent \( i \) at period \( t \) is given by the following equation:

\begin{equation} \label{eq:demand}
q_{i,t} = \frac{e^{\frac{\alpha_{i,t} - p_{i, t}}{\mu}}} {\sum_{j=1}^n e^{\frac{\alpha_{j,t} - p_{j, t}}{\mu}} + e^{\frac{\alpha_0}{\mu}}}
\end{equation}

where \( p_{i,t} \) denotes the agent's chosen price, \( \alpha_{i,t} \) represents the agent's vertical differentiation index, \( \mu \) corresponds to the horizontal differentiation parameter, and \( \alpha_0 \) indicates the inverse demand index.

We assume that production costs $c_t$ are identical across agents and grow at a rate of $w_t$:

\begin{equation} \label{eq:cost_growth}
    c_t = c_{t-1} \cdot (1 + \omega_t)
\end{equation}

where we assume that $\alpha_{i,t}$ grows at the same rate to ensure market persistence. 

Additionally, we assume that the price increase $\omega_t$ is modeled as a stochastic variable:

\begin{equation} \label{eq:inf_env}
    \omega_t = 
    \begin{cases}
        \bar{\omega}_i & \text{with probability $\rho$,} \\
        0 & \text{with probability $1 - \rho$.}
    \end{cases}
\end{equation}

where $\bar{\omega}_i$ denotes the observed monthly value of an actual inflation time series (see Section \ref{section:setup} for a detailed description).

Finally, the proposed Bertrand function with inflation is defined as:

\begin{equation} \label{eq:demmand_env}
q_{i,t} = \frac{e^{\frac{\alpha_{i,t} - p_{i, t}}{\lambda_t \cdot \mu}}} {\sum_{j=1}^n e^{\frac{\alpha_{j,t} - p_{j, t}}{\lambda_t \cdot \mu}} + e^{\frac{\alpha_0}{\lambda_t \cdot \mu}}}
\end{equation}

where $\lambda_t = \lambda_{t-1} \cdot (1 + \omega_t)$ captures the evolution of the market price index over time.

\subsection{Markov Decision Process} \label{mdp}

The proposed Markov Decision Process (MDP) models the decision-making dynamics of agents within a duopoly, where each agent determines its pricing strategy based on observed costs and historical market data. The agents' rewards are defined as the economic returns received by the company, adjusted for inflation, and can be expressed as

\begin{equation} \label{eq:revenue}
    R_{i,t} = \frac{(p_{i,t} - c_t) \cdot q_{i,t}}{\lambda_t}
\end{equation}

Following the methodology introduced by Calvano \cite{calvano}, the agents' actions are defined within a discrete action space of dimension $m$, where each value is uniformly distributed between $\eta_{min}$ and $\eta_{max}$

\begin{equation} \label{eq:k_actions}
    \eta_{i,t} = \eta_{min} + a_{i,t} \cdot \frac{\eta_{max} - \eta_{min}}{m-1}
\end{equation}

However, in contrast to this approach, actions in our model are interpreted as the target percentage profit margin \textit{over costs} set by each agent. This margin is then converted into the pricing policy employed by the agent as follows:

\begin{equation} \label{eq:actions}
    p_{i,t} = c_t \cdot (1 + \eta_{i,t})
\end{equation}

In each time step, agents make pricing decisions based on their observations of the current production cost $c_t$. Furthermore, as suggested in previous literature, agents are endowed with memory, allowing them to access both the costs and actions taken in the past periods $k$. This historical information is represented as:

\begin{equation}
    S_t =  \{ c_t \} \cup \left\{ \frac{P_{t-k} - c_{t-k}}{c_{t-k}}, ..., \frac{P_{t-1} - c_{t-1}}{c_{t-1}} \right\} \cup \{ c_{t-k}, ..., c_{t-1} \}
\end{equation}

\begin{figure}[h]
  \centering
  \includegraphics[width=16cm]{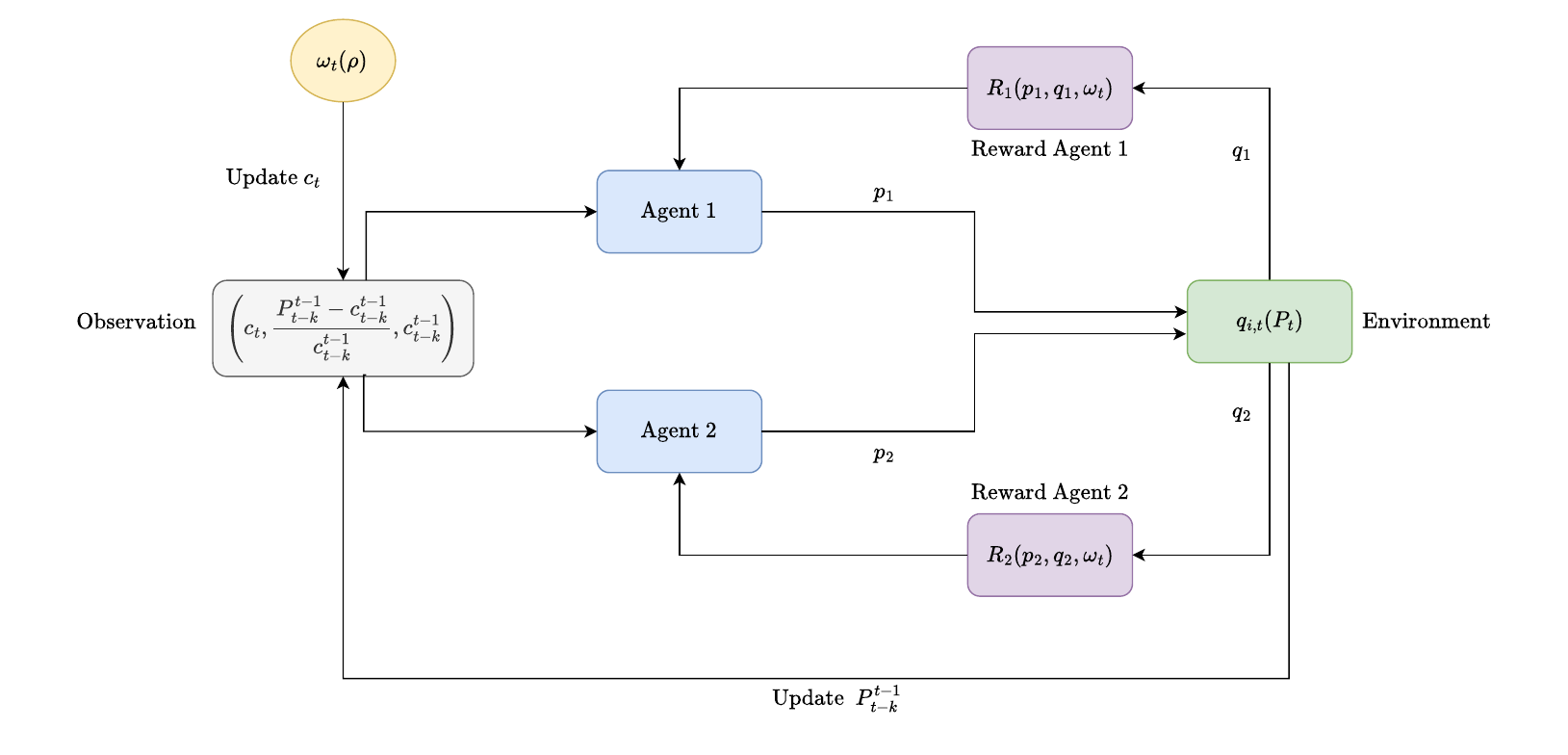}
  \caption{Proposed Markov Decision Process representing Bertrand Competition with increasing costs of production.}\label{fig:mdp}
\end{figure}

The general structure of the proposed MDP is illustrated in Figure \ref{fig:mdp}. In each period, agents use their available information to optimize their pricing decisions. These prices, in turn, interact with the market demand function, which determines the quantity allocated to each agent and, consequently, their resulting economic profit. Finally, updated prices are incorporated into historical price records, influencing future decisions.

\subsection{Agent} \label{section:agent}

\subsubsection{Deep Q-Network} \label{section:dqn}

We propose utilizing the Deep Q-Network (DQN) algorithm \cite{mnih2013playing} as the core reinforcement learning approach in our study. DQN, introduced by Mnih et al. in their seminal work Playing Atari with Deep Reinforcement Learning (2013) \cite{mnih2013playing}, represents a significant breakthrough in reinforcement learning, enabling the approximation of the optimal value function \(Q^*(s, a)\). This function predicts the expected cumulative reward when taking action \(a\) in state \(s\), guided by the Bellman equation:

\begin{equation}
    Q^*(s, a) = \mathbb{E}\left[R_t + \gamma \max_{a'}Q^*(s', a') \mid s, a\right]
\end{equation}

Here, \(R_t\) denotes the immediate reward, \(s'\) represents the next state, \(\gamma\) is the discount factor, and the expectation \(\mathbb{E}\) takes over the possible transitions of the environment.

DQN leverages a deep neural network to approximate \(Q^*\), employing techniques such as experience replay and a target network to stabilize training. These strategies address challenges such as temporal correlations and the non-stationarity inherent in reinforcement learning datasets. Specifically, the replay buffer mitigates overfitting by allowing the network to learn from a diverse set of experiences, while the target network decouples the updates, reducing oscillations in the learning process.

The algorithm's ability to handle high-dimensional, continuous state spaces and generalize from large datasets has established it as a cornerstone in deep reinforcement learning. However, limitations such as overestimation bias and the sensitivity to hyperparameter tuning remain critical concerns. These issues are particularly relevant in our study, where careful optimization is required to ensure reliable convergence in dynamic pricing scenarios.

Compared to traditional Q-learning methods, DQN has demonstrated superior performance in complex tasks, paving the way for more advanced approaches in deep reinforcement learning. Algorithm \ref{algo:dqn} provides the pseudocode of the DQN algorithm, highlighting the use of the \textit{Epsilon-greedy} strategy to address the exploration-exploitation trade-off. This implementation framework directly supports our objective of studying emergent cooperative behaviors in algorithmic pricing models.

\begin{algorithm}[H]
\SetAlgoLined
Initialize replay memory $\mathcal{D}$ to capacity $N$\;
Initialize action-value function $Q$ with random weights\;
\For{episode = 1, M}{
    Initialize sequence $s_1 = \{x_1\}$ and preprocessed sequence $\phi_1 = \phi(s_1)$\;
    \For{$t = 1, T$}{
        With probability $\epsilon$, select a random action $a_t$\;
        Select $a_t = \max_{a} Q^*(\phi(s_t), a; \theta)$\;
        Execute action $a_t$ in the emulator and observe reward $r_t$ and image $x_{t+1}$\;
        Set $s_{t+1} = s_t, a_t, x_{t+1}$ and preprocess $\phi_{t+1} = \phi(s_{t+1})$\;
        Store transition $(\phi_t, a_t, r_t, \phi_{t+1})$ in $\mathcal{D}$\;
        Sample a random minibatch of transitions $(\phi_j, a_j, r_j, \phi_{j+1})$ from $\mathcal{D}$\;
        Set $
        y_j =
        \begin{cases} 
        r_j & \text{for terminal } \phi_{j+1} \\
        r_j + \gamma \max_{a'} Q(\phi_{j+1}, a'; \theta) & \text{for non-terminal } \phi_{j+1}
        \end{cases}
        $
        
        Perform a gradient descent step on $(y_j - Q(\phi_j, a_j; \theta))^2$\;
    }
}
\caption{Deep Q-Learning Algorithm \cite{mnih2013playing}}\label{algo:dqn}
\end{algorithm}

\subsubsection{Epsilon Greedy}

Following the methodology proposed by Mnih et al. \cite{mnih2013playing}, we adopt the Epsilon-Greedy algorithm as an off-policy strategy to balance the exploration-exploitation trade-off.

\begin{equation}
a_t = 
\begin{cases} 
\text{random action}, & \text{with probability } \epsilon_t \\
\underset{a}{\mathrm{argmax}} \, Q(s_t, a), & \text{with probability } 1 - \epsilon_t
\end{cases}
\end{equation}

Here, the exploration rate is defined as $\epsilon_t = e^{-\beta t}$, where $\beta > 0$ is the exploration decay parameter. This exponential decay encourages firms to engage in exploratory actions during the early phases of the experiment, gradually transitioning toward actions with the highest estimated return as the experiment progresses.

\subsubsection{Neural Network Architecture}

In the study of \textit{Algorithmic Collusion} using reinforcement learning agents, the implementation of a neural network, denoted as \(\phi_{\theta}\), plays a pivotal role. As described in Section \ref{section:dqn}, this network is designed to process the state \(s_t\) and approximate the value function \(Q(s_t,a)\), formalized mathematically as:

\begin{equation}
\phi_{\theta} (s_t) \rightarrow Q(s_t,a)
\end{equation}

To achieve this, we propose a variant of the original algorithm \cite{mnih2013playing}, in which each agent utilizes a Multi-Layer Perceptron (MLP) composed of three hidden layers. The ReLU activation function is applied between layers due to its proven effectiveness in mitigating the vanishing gradient problem, a critical challenge in deep learning. Prior to feeding the state \(s_t\) into the neural network, a \textit{flatten} operation is performed to transform the state components into a format suitable for MLP processing.

The flattened components are then concatenated into a vector of dimensions \(N \times k + k + 1\), which undergoes normalization using a moving average with a window size of \(1/\rho\). This normalization step is essential for addressing non-stationarity issues commonly encountered in temporal data. A detailed graphical representation of the implemented neural network architecture is provided in Figure \ref{fig:architecture}, illustrating the relationships between key components:

\begin{figure}[h]
  \centering
  \includegraphics[width=12cm]{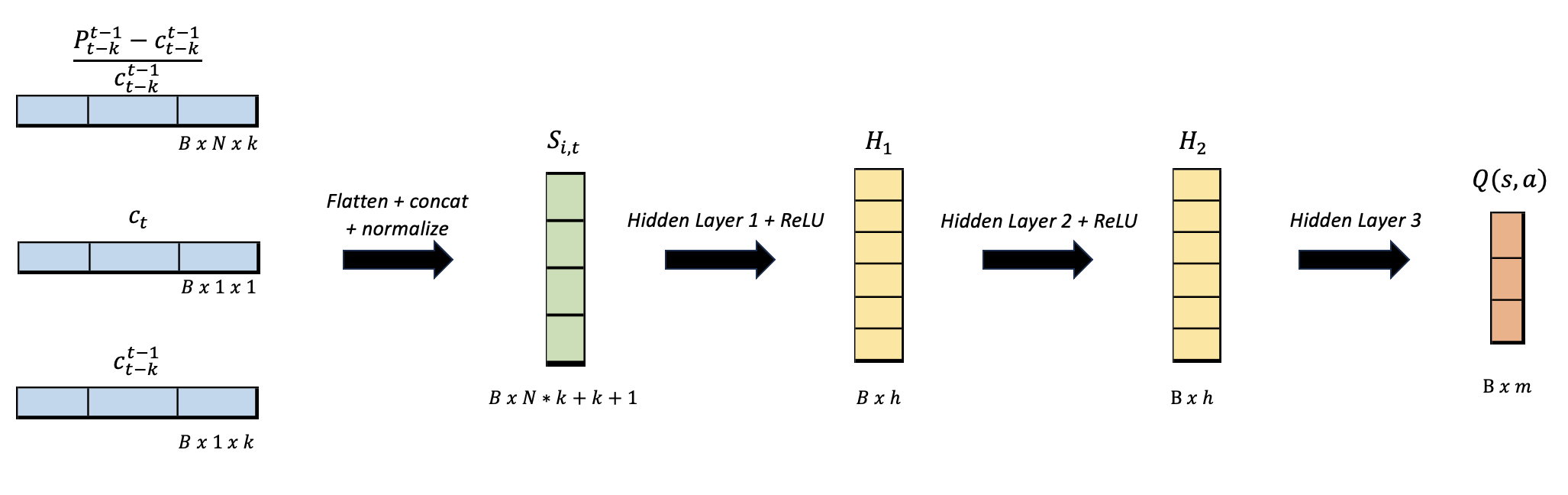}
  \caption{Neural Network Architecture for Agents.}\label{fig:architecture}
\end{figure}

In this architecture, \(B\) denotes the size of each batch sampled from the replay buffer, \(h\) represents the number of neurons in the hidden layers, \(m\) specifies the number of available actions, and \(H_i\) corresponds to the hidden state produced after processing data through the \(i\)-th linear layer. This design ensures efficient representation learning while addressing challenges inherent to algorithmic pricing with non-stationary costs.

\section{Experimental results}

\subsection{Experimental setup} \label{section:setup}

\subsubsection{Computation of Inflation}

To ensure realistic experimental conditions, we utilized inflation series from 10 different countries spanning the period 2000–2015 using World Bank Data \cite{ha2021one} (see Appendix \ref{appendix:inflation_series} for a brief description of the selected series). These series were used as inputs in our simulations to compute $\omega_t$, following the procedure outlined in the pseudocode described in Algorithm~\ref{algo: inflation}.

\begin{algorithm}
\caption{Inflation computation}
\SetAlgoLined
\For{each experimental run}{
    Randomly sample an inflation series $W_j$ \;
    Initialize series index $i = 0$ \;
    
    \For{each time step $t$}{
        Sample $l$ from a uniform distribution over [0,1] \;
        
        \eIf{$l < \rho$}{
            $\omega_t \gets \text{Draw from position $i$ of $W_j$}$ \;
            $i \gets i + 1$ \;
        }{
            $\omega_t \gets 0$ \;
        }
    }
}
\label{algo: inflation}
\end{algorithm}

\subsubsection{Base parameters} \label{section:base_params}

This study aims to assess the impact of incorporating inflation on market competitiveness under algorithmic collusion. To this end, a comparative static analysis is performed by varying the frequency of inflation shocks, using the base parameter configuration (see Apppendix \ref{appendix:base_params}). The analysis focuses on two scenarios: Algorithmic Competition without Inflation (\(\rho = 0.000\)) and Algorithmic Competition with Inflation (\(\rho = 0.001\)).

\subsubsection{Profits}

To assess the level of market competitiveness, we adopt the methodology proposed by \cite{calvano}, which defines \(\Delta_t\) as a measure of profitability relative to the Nash and Monopolistic equilibria:
\begin{equation} \label{eq:Delta_t}
    \Delta_t = \frac{\bar{R_t} - R^N_t}{R^M_t - R^N_t}
\end{equation}

In this equation, \(\bar{R}_t\) denotes the average profitability achieved by the agents at time \(t\), while \(R^N_t\) and \(R^M_t\) represent the profitabilities corresponding to the Nash and Monopolistic equilibria, respectively (see Appendix \ref{appendix:monopoly_nash}).

Since agents' actions are tied to costs (see Equation \ref{eq:actions}), and costs may grow at a different rate than the Nash and Monopolistic equilibria, the values of \(\Delta_t\) are subject to an inherent bias arising from discrepancies between inflation growth rates and those of the Nash and Monopolistic equilibria. To mitigate this issue, we define a inflation aware measure of profitability \(\nabla_t\) as follows:
\begin{equation} \label{eq:nabla}
    \nabla_t = \frac{\bar{R}_t - R^{N_F}_t}{R^{M_F}_t - R^{N_F}_t}
\end{equation}

Here, \(R^{N_F}_t\) and \(R^{M_F}_t\) denote the Nash and Monopolistic profitabilities adjusted to account for inflation-equivalent growth. By using \(\nabla_t\), it becomes possible to detect cooperative strategies while controlling for differences in growth between the series. For a complete understanding of $\nabla$ and how it relates to $\Delta$ see Appendix \ref{appendix:nabla}

Within this framework, lower values of \(\Delta_t\) and \(\nabla_t\) indicate profitability levels consistent with a competitive market. Conversely, higher values suggest the presence of a non-competitive equilibrium, which could potentially harm consumer welfare.

\subsubsection{Statistical Robustness}

To ensure statistical robustnesss, each experiment is repeated \(N = 50\) times. This repetition introduces variability across experiments based on the following factors:

\begin{itemize}
    \item The inflation series used to extract \(\pi_t\),
    \item The initialization weights assigned to each agent’s network, and
    \item The strategy selected by the \textit{epsilon-greedy} algorithm during the exploration phase.
\end{itemize}

The level of competitiveness across experiments is quantified using the statistic \(\mu\), defined as:

\begin{equation}
    \mu = \frac{1}{N} \frac{1}{T} \sum_{i=1}^{N} \sum_{t=1}^{T} \nabla_{i, t}
\end{equation}

where \(T\) represents the total number of timesteps.

To assess the effect of inflation, we compute the \textit{Cohen's d} statistic \cite{cohen2013statistical}, which measures the standardized difference between the means of two paired groups. It is calculated as:

\begin{equation}
    d = \frac{\overline{D}}{s_D}
\end{equation}

where \(\overline{D}\) denotes the mean difference between the two groups, and \(s_D\) is the standard deviation. Following the methodology outlined by \cite{cohen2013statistical}, the effect size is categorized according to the intervals in Table \ref{tab:cohen}.

\begin{table}
  \centering
  \caption{Intervals of $d$ and their corresponding effect size.}
  \label{tab:cohen}
  \begin{tabular}{cc}
    \toprule
    \textbf{$d$} & Effect Size \\
    \midrule
    $d < 0.01$                & Negligible               \\ 
    $0.01 \leq d < 0.20$      & Very small               \\ 
    $0.20 \leq d < 0.50$      & Small                    \\ 
    $0.50 \leq d < 0.80$      & Medium                   \\ 
    $0.80 \leq d < 1.20$      & Large                    \\ 
    $1.20 \leq d < 2.00$      & Very large               \\
    $d \geq 2.00$             & Huge                     \\ 
    \bottomrule
  \end{tabular}
\end{table}

Finally, we perform a comparative analysis using an independent two-sample \textit{t}-test to evaluate whether the differences between the in-sample and out-of-sample configurations are statistically significant. These configurations are defined as follows:

\begin{itemize}
    \item \textbf{In-sample} ($I$): Agents are jointly trained and evaluated over the entire duration of the experiment, starting from uninitialized parameters.
    \item \textbf{Out-of-sample} ($O$): Agents are first trained using the baseline configuration described in Section~\ref{section:base_params}. A random pair of agents is then selected, their parameters are fixed (i.e., gradients are frozen), and they are evaluated in a previously unseen market environment for 50{,}000 timesteps.
\end{itemize}

\subsubsection{Reward-punishment scheme}

As stated by previous works \cite{calvano, lepore, klein, abada}, the mere presence of monopolistic profits does not provide definitive evidence that the equilibrium arises from collusive behavior. Since economic collusion relies on a sophisticated interplay of incentives and punishments to sustain prices consistently above the competitive threshold—thereby securing higher economic profitability—it is crucial to test for the existence of punishment mechanisms. Such analysis is essential to rigorously differentiate true collusion from irregularities in strategic decision-making.

In line with the above and based on the methodologies proposed, the objective of this experiment is to evaluate whether the supra-competitive equilibrium reached originates from cooperative actions. To achieve this, an evaluation of the existence of punishment strategies is suggested by inducing deviations in one agent and subsequently analyzing the counterpart's response. Specifically, a condition is proposed where one of the agents adopts the price corresponding to the Nash equilibrium at $\bar t$, expressed by the following equation:
\begin{equation} \label{eq:deviate}
    p_{0,\bar t} = p_{N,\bar t}
\end{equation}

In this manner, a cooperation supporting case is evaluated as those scenarios where the agent responds to the deviation with a reduction in its pricing policy within the 5 subsequent timesteps after the deviation. Conversely, a negative case is defined as those instances where the agent makes no adjustments to its actions within the same time window. Finally, considering that the desired outcome is to evaluate the existence of punishment strategies once the agents have learned the most optimal strategy, it is proposed to enforce the agents' deviation at $\bar t = 350{,}000$.

\subsection{Main results} \label{section:results}

\begin{figure}[h]
  \centering
  \begin{subfigure}[b]{0.45\textwidth}
    \centering
    \includegraphics[width=\linewidth]{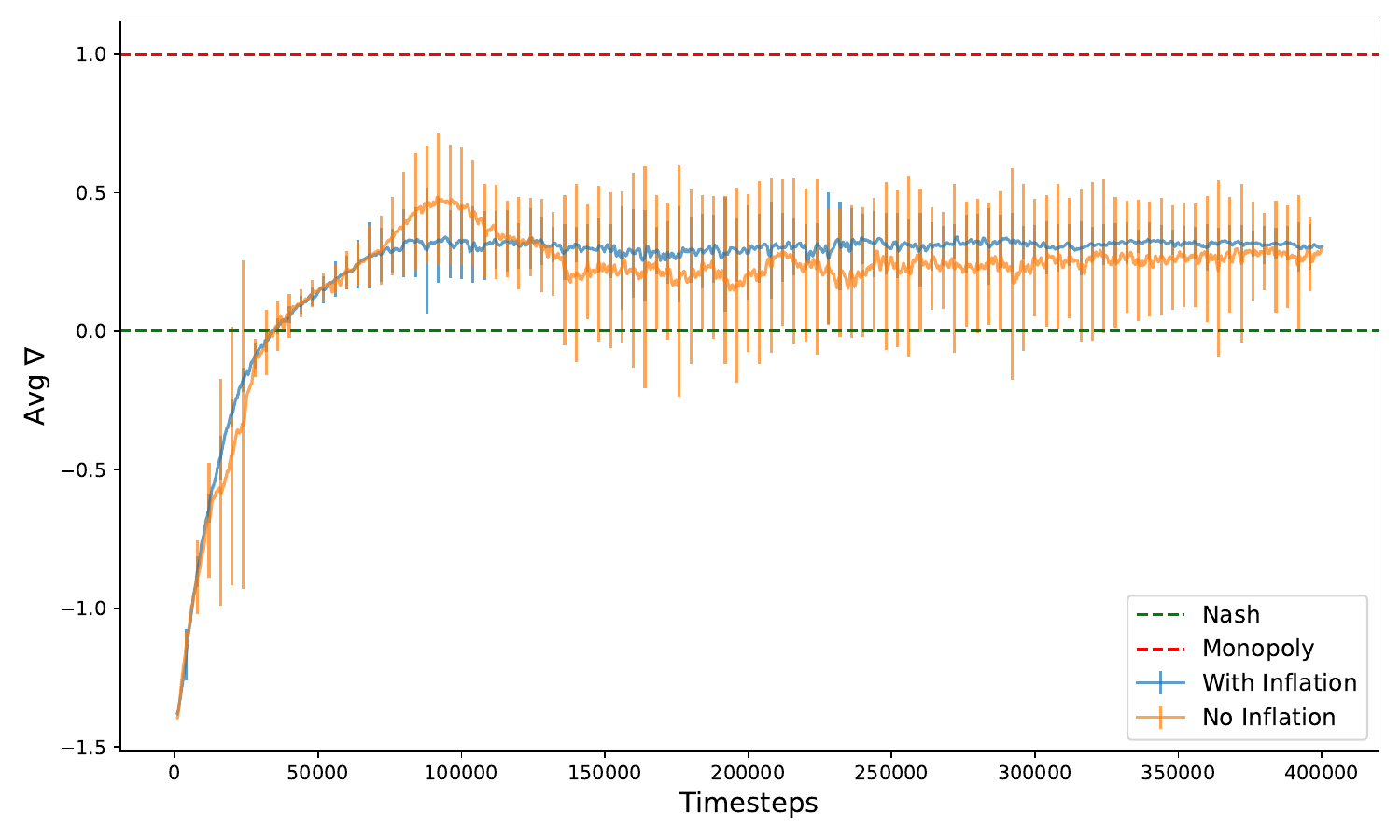}
    \subcaption{Profits evolution across experiments using in-sample data.}
    \label{fig:train}
  \end{subfigure}
  \hfill
  \begin{subfigure}[b]{0.45\textwidth}
    \centering
    \includegraphics[width=\linewidth]{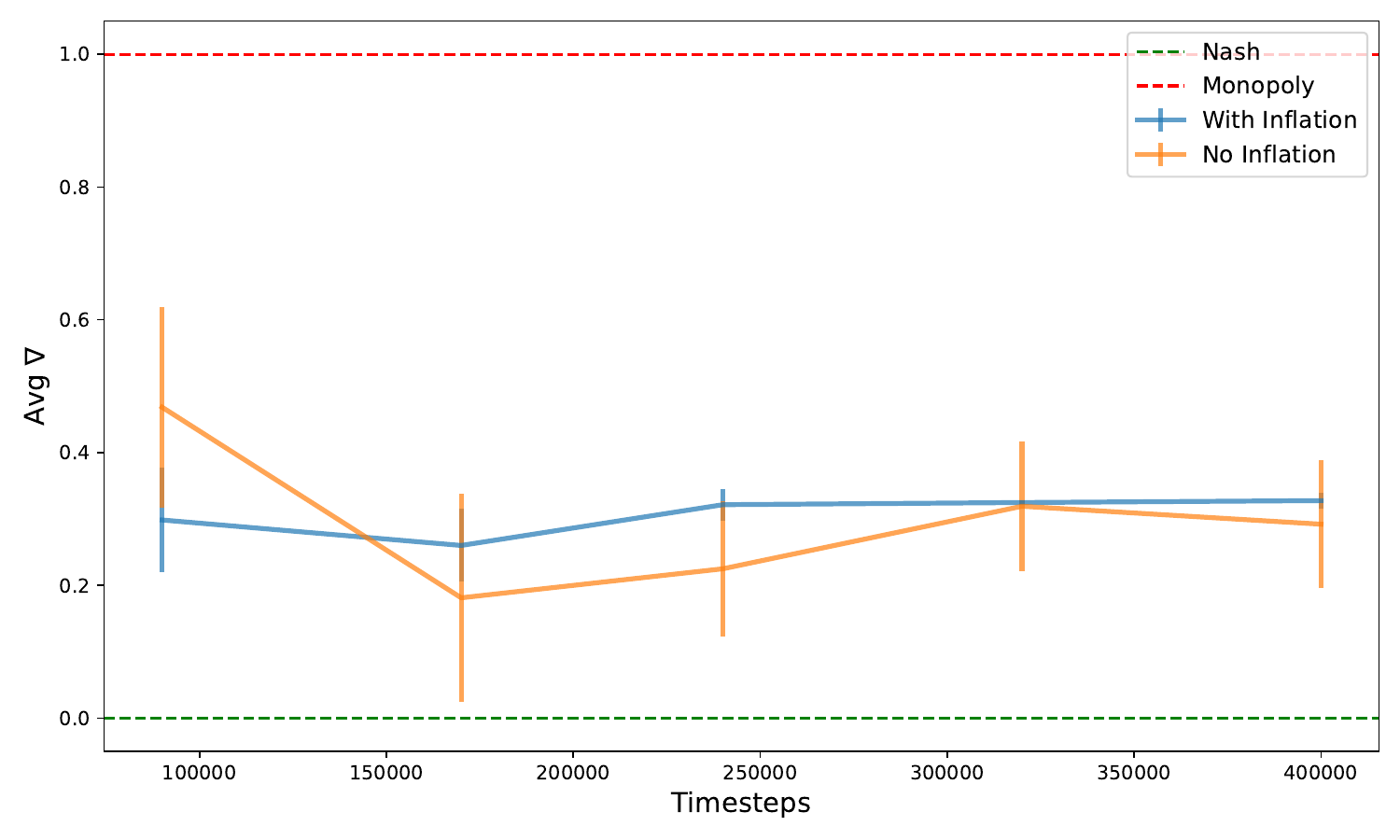}
    \subcaption{Profits evolution across experiments using out-of-sample data.}
    \label{fig:test}
  \end{subfigure}
  \caption{Comparison of profit evolution across experimental settings with and without inflation. The blue and orange lines depict the profitability levels $\nabla$ under inflationary and non-inflationary conditions, respectively. Given the variance in prices set by individual agents, the figure shows a 1000-timestep moving average along with its corresponding confidence interval. For reference, the Nash and Monopoly equilibrium outcomes are also included, represented by the green and red lines, respectively.}
  \label{fig:combined}
\end{figure}

In Figure~\ref{fig:combined}, we present the evolution of profits across experimental settings, comparing scenarios with and without inflation. Panel \ref{fig:train} reports in-sample performance, while panel \ref{fig:test} illustrates out-of-sample results. These results demonstrate that agents consistently attain profitability levels above the Nash equilibrium in all configurations. Furthermore, the experiments that incorporate inflation exhibit systematically higher average profits, suggesting that inflation contributes to a less competitive market environment.

Similarly, and notably, out-of-sample configurations exhibit lower competitiveness compared to in-sample settings, establishing the inflation-inclusive out-of-sample configuration as the least competitive. Table \ref{tab:main_results} summarizes these findings.  

\begin{table}[t]
    \centering
    \caption{Agents profits $\mu$ across experiments}
    \label{tab:main_results}
    \begin{tabular}{ccc}
        \toprule
        Configuration & In-sample & Out-of-sample \\
        \midrule
        No Inflation & $0.2746$ & $0.2920$ \\
        With Inflation & $0.3156$ & $0.3258$ \\
        \bottomrule
    \end{tabular}
\end{table}

While results show a clear distinction across configurations, the overall impact of configuration changes remains limited. Specifically, including or excluding inflation in the experiments has a marginal effect, similar to the differences observed between in-sample and out-of-sample evaluations (see Tables \ref{tab:effect_size} and \ref{tab:sample_comparison}).

\begin{table}[t]
  \label{tab:all}
  \centering
  \begin{minipage}{0.3\textwidth}
    \centering
    \caption{Inflation Effect Size}
    \label{tab:effect_size}
    \begin{tabular}{ccl}
      \toprule
      Configuration&$d$&Effect Size\\
      \midrule
      In-sample & 0.3272 & Small\\
      Out-of-sample & 0.3047 & Small\\
      \bottomrule
    \end{tabular}
  \end{minipage}
  \hspace{1cm}
  \begin{minipage}{0.3\textwidth}
    \centering
    \caption{Sample comparison}
    \label{tab:sample_comparison}
    \begin{tabular}{cl}
      \toprule
      Configuration&$p(\mu_I = \mu_O)$\\
      \midrule
      No Inflation & 0.5735\\
      With Inflation & 0.3937 \\
      \bottomrule
    \end{tabular}
    \end{minipage}
\end{table}

\subsubsection{Actions}

\begin{figure}[h]
  \centering
  \begin{subfigure}{0.45\textwidth}
      \centering
      \includegraphics[width=\linewidth]{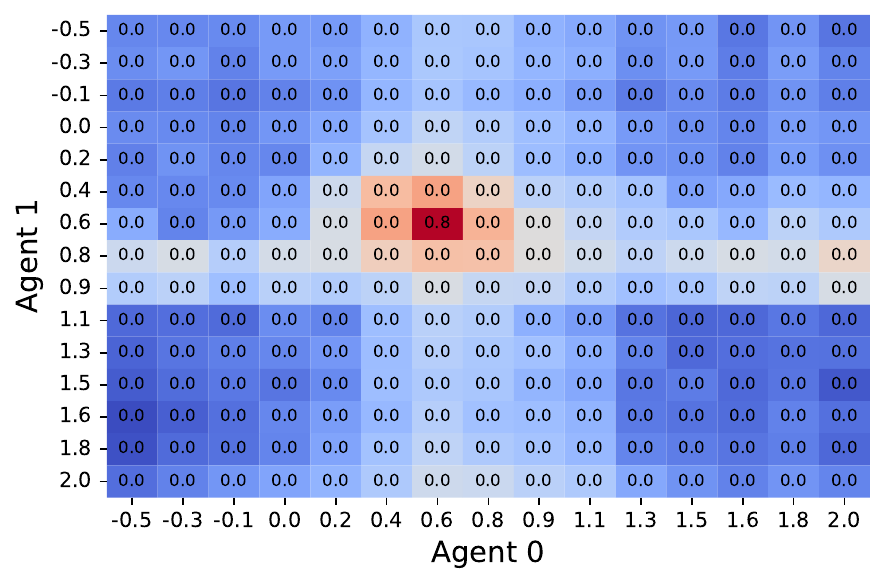}
      \caption{No Inflation — In-sample} 
      \label{fig:heatmap_no_inflation_train}
  \end{subfigure}
  \hfill
  \begin{subfigure}{0.45\textwidth}
      \centering
      \includegraphics[width=\linewidth]{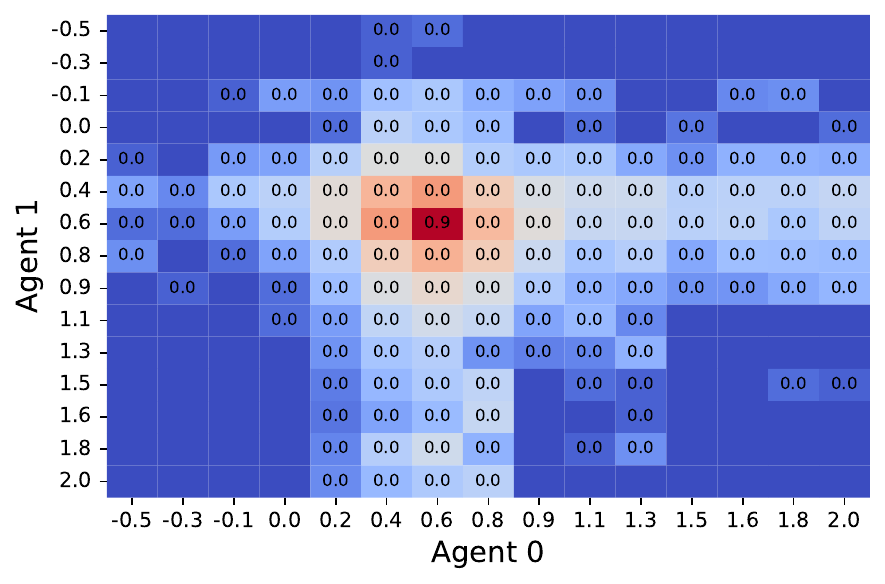}
      \caption{Inflation — In-sample} 
      \label{fig:heatmap_inflation_train}
  \end{subfigure}

  \vspace{0.5cm}

  \begin{subfigure}{0.45\textwidth}
      \centering
      \includegraphics[width=\linewidth]{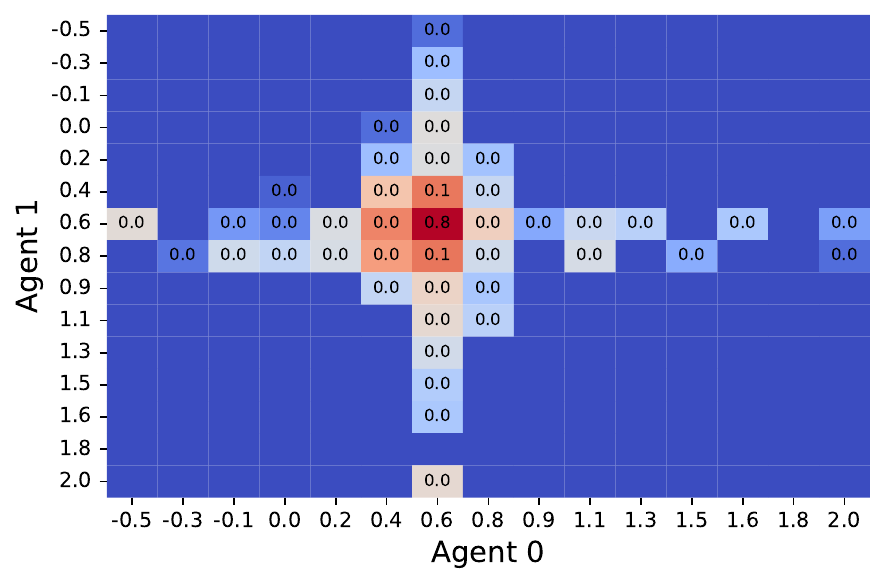}
      \caption{No Inflation — Out-of-sample} 
      \label{fig:heatmap_no_inflation_test}
  \end{subfigure}
  \hfill
  \begin{subfigure}{0.45\textwidth}
      \centering
      \includegraphics[width=\linewidth]{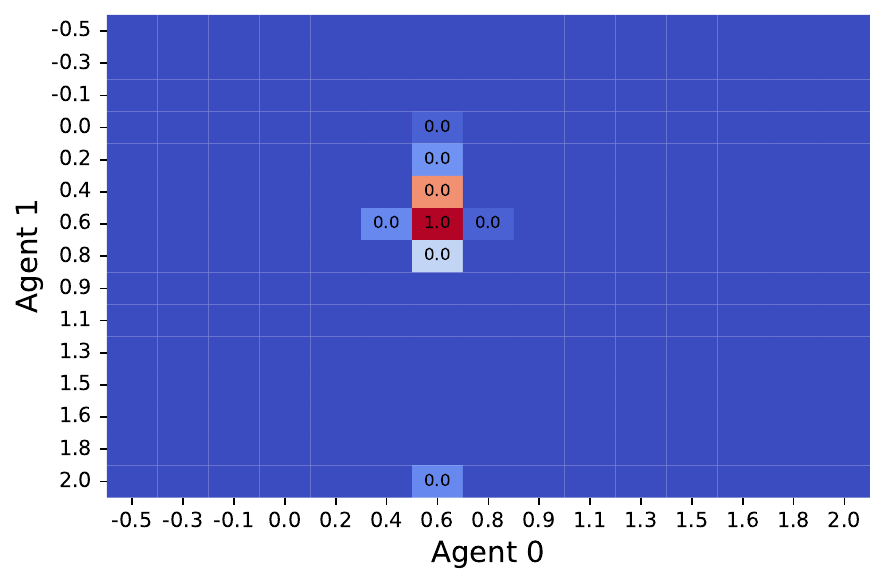}
      \caption{Inflation — Out-of-sample} 
      \label{fig:heatmap_inflation_test}
  \end{subfigure}

\caption{Prices over costs (\%) for experiments with and without inflation using both in-sample and out-of-sample data. For each configuration, the empirical joint distribution of action pairs is visualized as a percentage heatmap using a logarithmic color scale. The x-axis corresponds to the pricing decisions made by Agent 0, while the y-axis represents those of Agent 1. In Figure~\ref{fig:heatmap_no_inflation_train}, for instance, 80\% of the observed timesteps show both agents selecting prices that are 60\% above their respective costs.}
\label{fig:combined_heatmaps}
\end{figure}

To complement the previous analysis, it is valuable to examine the actions taken by agents when competing through prices. As outlined in Section \ref{mdp}, actions are defined as price markups over costs, within an interval of $m$ possible values ranging from $\eta_{min}$ to $\eta_{max}$. Figures \ref{fig:heatmap_no_inflation_train} and  \ref{fig:heatmap_inflation_train} present heatmaps of agent actions in configurations with and without inflation using in-sample data, while Figures \ref{fig:heatmap_no_inflation_test} and  \ref{fig:heatmap_inflation_test} provide the same analysis for out-of-sample data.  

The results indicate that agents predominantly set their prices around $60\%$ above costs, a pattern that remains consistent across inflationary and non-inflationary conditions, as well as in both in-sample and out-of-sample scenarios. However, a notable pattern emerges: under inflationary conditions, agent prices exhibit a stronger tendency to cluster around this threshold, whereas in a stable price environment, their pricing decisions display greater dispersion.  

A comparable trend is observed when contrasting in-sample and out-of-sample data: agent pricing decisions tend to exhibit greater clustering in out-of-sample datasets, with the most pronounced clustering effect occurring in the scenario where inflation is present and agents operate in out-of-sample environments.

\subsubsection{Deviations and punishments} \label{section:deviation}

\begin{figure}[h]
  \centering
  \begin{subfigure}{0.47\textwidth}
      \centering
      \includegraphics[width=\linewidth]{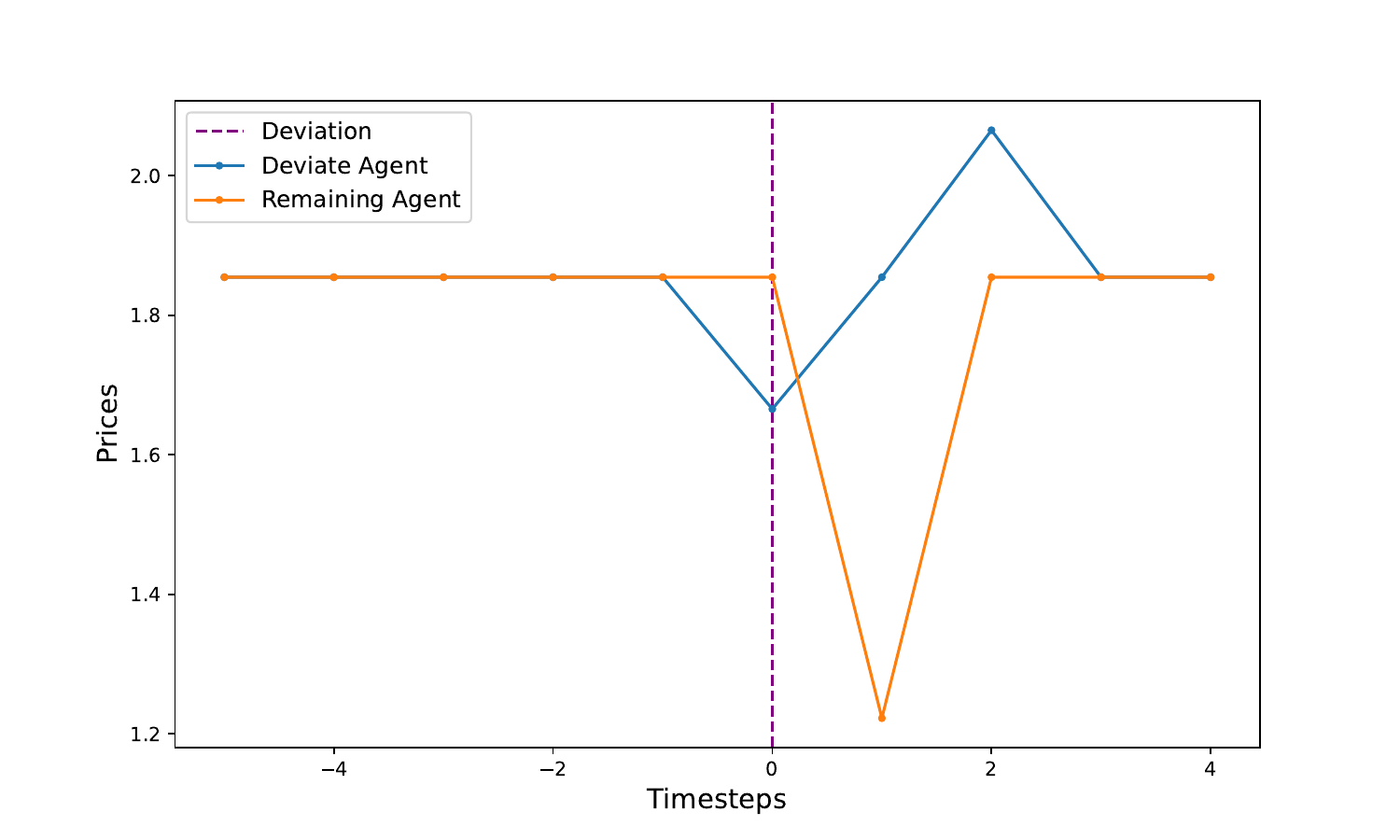}
      \caption{Sample with punishment}
  \end{subfigure}
  \hfill
  \begin{subfigure}{0.47\textwidth}
      \centering
      \includegraphics[width=\linewidth]{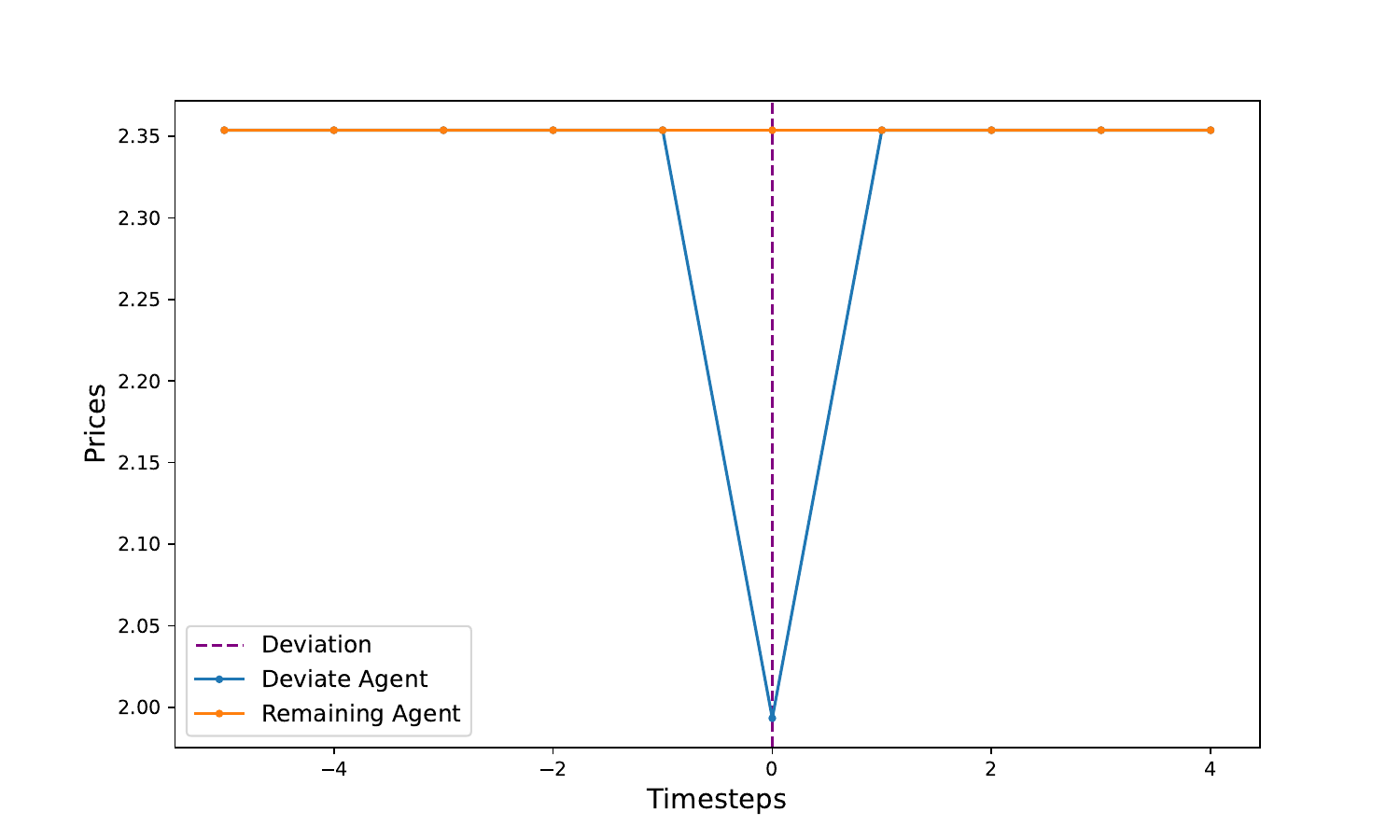}
      \caption{Sample with no punishment}
  \end{subfigure}
\caption{Illustrative outcomes of the reward-punishment experiments. The blue line denotes the agent subjected to the forced deviation, the orange line indicates the responding agent whose behavior is under evaluation, and the purple vertical line marks the timestep at which the deviation is introduced.}
\label{fig:punish}
\end{figure}

The results presented in Table \ref{tab:deviate} indicate that, among the 50 experiments conducted, only two exhibit a discernible shift in pricing behavior—likely as a form of punishment, reflected in a sharp reduction in price levels. Conversely, the remaining cases display price rigidity, with agents failing to adjust their strategies despite competitors' price deviations. This finding is particularly noteworthy, as the optimal strategy response to equilibrium deviations entails undercutting to preserve profitability.  The fact that most experiments do not exhibit such behavior suggests that reward-punishment schemes are not effectively learned under this experimental setup, raising concerns about the context-dependence and limited external validity of the mechanisms observed in \cite{calvano}. Figure \ref{fig:punish} illustrates representative instances where punitive strategies are either clearly manifested or entirely absent.

\begin{table}
  \centering
  \caption{Experiment results for punishment strategies}
  \label{tab:deviate}
  \begin{tabular}{ccl}
    \toprule
    Case&$N$&Proportion\\
    \midrule
    With punishment & 2 & 0.04\\
    No punishment & 48 & 0.96\\
  \bottomrule
\end{tabular}
\end{table}

\subsection{Analysis}

The results presented in Section \ref{section:results} exhibit the role of inflation in influencing market competitiveness. Specifically, the findings suggest that accounting for cost fluctuations drives agents to converge toward a less competitive equilibrium than scenarios where such variations are disregarded. Furthermore, analysis of actions reveals that inflation enhances agents' profitability and promotes greater stability in their pricing strategies.

These results could theoretically be interpreted as an emergent phenomenon wherein agents exploit price volatility to optimize their profitability. However, a more detailed examination challenges this hypothesis. As demonstrated in Section \ref{section:deviation}, the observed profitability gains cannot be attributed to cooperative behavior but rather to an exogenous, unobserved factor. Moreover, as discussed in \cite{abada}, the prevailing hypothesis suggests that agents do not consistently converge towards the optimal pricing strategy—setting prices marginally below those of their competitors—which ultimately fosters non-competitive market dynamics.

These findings carry significant implications for the existing literature. First, they suggest that prior studies may have underestimated the detrimental effects of algorithmic competition on market competitiveness by failing to account for price fluctuations. Additionally, as evidenced in the stock movement analysis, high-inflation environments may contribute to stabilizing agents' pricing policies, thereby amplifying their market power.

Although these results do not provide direct evidence of cooperative strategies among agents, they remain of considerable relevance to market regulators. Notably, autonomous agents can sustain profitability levels exceeding the competitive equilibrium, even under inflationary conditions. Furthermore, these dynamics could materialize in real-world markets, irrespective of explicit agent cooperation. As demonstrated in \cite{calvano}, despite an extended experimental horizon, agents exhibit anti-competitive behaviors in the early stages of the simulation. Specifically, assuming that 1,000 timesteps correspond to one month in real time—equivalent to approximately 1.38 actions per agent per hour—such behaviors could manifest as early as 33 months post-deployment. 

Consistent with previous analyses, the results indicate that configurations incorporating inflation lead to non-competitive outcomes at a faster rate compared to those with stable price levels. Table \ref{tab:times} presents the minimum number of steps required to attain supra-competitive economic returns under the in-sample configurations.

\begin{table}[h]
    \centering
    \caption{Average Minimum Steps to Achieve Supra-Competitive Profits}
    \label{tab:times}
    \begin{tabular}{cc}
        \toprule
        Configuration & Steps \\
        \midrule
        No Inflation & 34{,}712 \\
        With Inflation & 33{,}904 \\
        \bottomrule
    \end{tabular}
\end{table}

Finally, while these findings underscore the relevance of inflation in the study of algorithmic collusion, further investigation is required to assess its broader implications across diverse market environments. Appendix~\ref{appendix:sensitivity} provides a preliminary sensitivity analysis, examining variations in critical parameters: the number of agents ($N$), the learning rate ($lr$), the number of observed periods ($k$), the intertemporal discount factor ($\gamma$), and the inflation rate ($\rho$). Results show that market competitiveness is highly sensitive to these model settings. In particular, increasing the number of agents $N$ leads to a significant enhancement in competitive dynamics, as evidenced by a pronounced decrease in $\mu$. This outcome aligns with established economic theory, wherein a greater number of firms shifts the market structure from oligopoly toward perfect competition. Furthermore, the result is consistent with prior findings in~\cite{dawid2024deep, hettich2021algorithmic}, reinforcing the claim that increasing the number of competitors serves as an effective mechanism to mitigate the adverse effects of algorithmic collusion.

\section{Conclusions and Future work}

Our research investigates the potential adverse effects that algorithms may impose on society when utilized in critical decision-making contexts. Specifically, we focus on reinforcement learning-based algorithms and their capacity to adapt within dynamic environments. The objective of this work is to examine the hypothesis that, when applied to pricing decisions, these algorithms have the potential to maintain non-competitive equilibria, even amid fluctuations in production costs.

The findings partially support the research hypothesis. Introducing inflation into the experimental framework significantly impacts the profitability achieved by agents. Across both in-sample and out-of-sample settings, inflationary scenarios yielded higher profitability compared to their non-inflationary counterparts. Furthermore, agents operating in inflationary conditions exhibited greater coordination in their actions and reached non-competitive profit levels more rapidly. This suggests that inflation may undermine market competitiveness by facilitating implicit coordination among economic agents.

Conversely, the results do not provide conclusive evidence that agents develop effective punishment strategies to deter deviations from equilibrium. Only two of the fifty experiments conducted exhibited appropriate responses to equilibrium deviations. This contrasts with previous studies that report stronger reactive mechanisms. A plausible explanation lies in algorithmic differences: while this study employs Deep Q-Networks (DQN), prior research predominantly utilizes Q-learning, which may account for the observed variation in learned strategies.

A deeper investigation into how variability in production costs affects market competition dynamics reveals several promising research directions.  One approach is to expand the theoretical economic framework of Bertrand Competition to incorporate empirical market data. Another avenue is to integrate more advanced algorithms into the analysis to evaluate their influence on competitive dynamics. This is particularly relevant with the advent of autonomous agents powered by Large Language Models (LLMs), whose impact on market competitiveness needs more exploration. Lastly, the findings highlight the importance of developing regulatory policies that explicitly account for inflation in algorithmic competition to safeguard market competitiveness.

\bibliographystyle{unsrt}  
\bibliography{references}  

\section*{Appendices}

\appendix

\section{Inflation series} \label{appendix:inflation_series}

\begin{table}[h]
\centering
\caption{Summary: National Inflation Rates used on experiment}\label{tab:inflation_series}
\begin{tabular}{llllllll}
Country &  Average & Std & Min & 25\% &  50\% & 75\% & Max \\
\hline
Canada & 0.16 & 0.38 & -0.98 & -0.10 & 0.13 & 0.39 &
1.15 \\
China & 0.19 & 0.62 & -1.39 & -0.20 & 0.10 & 0.54 &
2.60 \\
France & 0.12 & 0.32 & -1.00 & -0.10 & 0.11 & 0.33 & 1.08 \\
Germany & 0.12	& 0.35 & -1.01 & -0.10 & 0.11 & 0.37 & 1.06 \\
Italy & 0.14 & 0.21 & -0.58 & 0.00 & 0.19 & 0.27 & 0.62 \\
Netherlands & 0.16 & 0.48 & -1.00 & -0.24 & 0.11 & 0.49 & 1.21 \\
Singapore & 0.13 & 0.50 & -1.61 & -0.20 & 0.10 & 0.44 & 2.01 \\
Sweden & 0.11 & 0.41 & -1.34 & -0.11 & 0.11 & 0.40 & 1.04 \\
Swiss  & 0.04	& 0.35	& -1.03	& -0.11	& 0.00	& 0.20	& 1.12 \\
United States & 0.18 & 0.38 & -1.97 & 0.00 & 0.19 & 0.41 & 1.21 \\ \hline
\end{tabular}

\end{table}

\section{Base Parameters} \label{appendix:base_params}

\begin{table}[h]
  \centering
  \caption{Base Parameters used on experiment.}
  \label{tab:base_params}
  \begin{tabular}{cc}
    \toprule
    Hyperparameter & Value \\ 
    \midrule
    Number of Agents (\(N\))        & 2  \\ 
    Number of Past Periods (\(k\))  & 1  \\ 
    Probability of Inflationary Shock (\(\rho\)) & 0.001 \\ 
    Learning Rate (\(lr\))          & 0.01 \\ 
    Discount Factor (\(\gamma\))    & 0.95 \\ 
    Minimum Cost Variation (\(\eta_{min}\)) & -0.5 \\ 
    Maximum Cost Variation (\(\eta_{max}\)) & 2.0 \\ 
    Number of Actions (\(m\))       & 15 \\ 
    Optimizer                       & Adam \cite{adam} \\ 
    Initial Production Cost (\(c_{t=0}\)) & 1 \\ 
    Initial Vertical Differentiation (\(\alpha_{t=0}\)) & 1 \\ 
    Inverse Aggregate Demand Index (\(\alpha_0\)) & 0 \\ 
    Horizontal Differentiation (\(\mu\)) & 0.25 \\ 
    Number of Neurons per Layer (\(h\)) & 256 \\ 
    Hidden Layers                   & 2 \\ 
    Epsilon Decay (\(\beta\))       & 0.00005 \\ 
    Batch Size (\(B\))              & 256 \\ 
    Replay Buffer Size              & 20,000 \\ 
    Episodes                        & 1 \\ 
    Timesteps                       & 400,000 \\ 
    Gradient Steps                  & 1 \\ 
    Target Update Steps             & 200 \\ 
    \bottomrule
  \end{tabular}
\end{table}

\section{Monopoly and Nash Equilibrium Computation} \label{appendix:monopoly_nash}

\subsection{Nash Equilibrium}

To compute the Nash equilibrium in a market with differentiated products, it is necessary to derive the reaction functions of the competing firms. These functions are obtained by differentiating the profit function with respect to price, using the demand function given in equation \ref{eq:demand} as a reference:

\begin{equation}
\frac{\partial R_i}{\partial p_i} = \frac{\partial}{\partial p_i} (p_i - c_t) \cdot q_{i,t}    
\end{equation}

\begin{equation}
\frac{\partial R_i}{\partial p_i} = \frac{\partial}{\partial p_i} (p_i - c_t) \cdot  \frac{e^{\frac{\alpha_{i,t} - p_{i, t}}{\mu}}} {\sum_{j=1}^n e^{\frac{\alpha_{j,t} - p_{j, t}}{\mu}} + e^{\frac{\alpha_0}{\mu}}}    
\end{equation}

The Nash equilibrium is determined by simultaneously solving the system of reaction functions, which requires equating the profit-maximizing conditions for all firms:

\begin{equation} \label{eq:nash}
    \frac{\partial R_i}{\partial p_i} - \frac{\partial R_j}{\partial p_j} = 0
\end{equation}

Here, \( R_i \) denotes the profit function of firm \( i \), and the remaining variables are defined in the main text. Due to the lack of a closed-form solution, equilibrium prices are computed numerically using standard optimization algorithms.

\subsection{Monopoly Equilibrium}

In a scenario where firms engage in collusion, the market can be modeled as a monopoly. Under this assumption, the profit maximization problem simplifies, and the demand function from equation \ref{eq:demand} reduces to:

\begin{equation} \label{eq:monopoly}
q_{i,t} = \frac{e^{\frac{\alpha_i - p_{i, t}}{\mu}}} {e^{\frac{\alpha_i - p_{i, t}}{\mu}} + e^{\frac{\alpha_0}{\mu}}}
\end{equation}

The monopolistic equilibrium is obtained by maximizing total revenue with respect to price:

\begin{equation}
    \max_{p_{i,t}} R_{i} = \sum_t (p_{i,t} - c_{t}) \cdot q_{i,t}
\end{equation}

\begin{equation}
    \max_{p_{i,t}} R_{i} = \sum_t (p_{i,t} - c_{t}) \cdot \frac{e^{\frac{\alpha_i - p_{i, t}}{\mu}}} {e^{\frac{\alpha_i - p_{i, t}}{\mu}} + e^{\frac{\alpha_0}{\mu}}}
\end{equation}

As in the Nash case, \( R_i \) denotes firm \( i \)'s profit, and all parameters follow definitions provided earlier. Given the nonlinearity of the objective function, optimal prices are obtained using numerical optimization methods.

\section{Decomposition of $\Delta_t$: Cooperation and Inflation Effects} \label{appendix:nabla}

From Equation \ref{eq:Delta_t}, $\Delta_t$ is determined by three key factors:

\begin{itemize}
    \item Agents' Prices $P(c_t)$
    \item Nash Equilibrium Price $p^N$
    \item Monopoly Price $p^M$
\end{itemize}

To ensure a consistent analysis of $\Delta_t$ over time, these three elements should exhibit proportional variation, thereby preventing distortions where one factor disproportionately offsets another. However, empirical observations based on the parameter set defined in Table \ref{tab:base_params} indicate that this condition was not met. The following discussion introduces a methodological approach to address this imbalance through a refined measurement metric, $\nabla_t$.

Considering the forced equilibria defined by the following equations:

\begin{equation} \label{eq:new_nash}
    p_{t+1}^{N_F} = p_t^{N_F} \cdot (1 + \pi)
\end{equation}

\begin{equation} \label{eq:new_monopoly}
    p_{t+1}^{M_F} = p_t^{M_F} \cdot (1 + \pi)
\end{equation}

We can use Equation \ref{eq:demmand_env} to compute the corresponding economic profits as follows:

\begin{equation}
    R_t^{N_F} = (p_t^{N_F} - c_t) \cdot q(p_t^{N_F})
\end{equation}

\begin{equation}
    R_t^{M_F} = (p_t^{M_F} - c_t) \cdot q(p_t^{M_F})
\end{equation}

Using the above expressions, we can express $R_t^N$ in terms of $R_t^{N_F}$:

\begin{equation}
    R_t^N = R_t^{N_F} + R_t^{\pi}
\end{equation}

where $R_t^{\pi}$ represents the profit variations driven by price level changes $\pi_t$.

By leveraging these equations, $\Delta_t$ can be reformulated as:

$$\Delta_t  = \frac{\bar{R}_t - R^{N}_t}{R^{M}_t - R^{N}_t}$$

$$\Delta_t \cdot \frac{R^{M}_t - R^{N}_t}{R^{FM}_t - R^{FN}_t} = \frac{\bar{R}_t - R^{N}_t}{R^{M}_t - R^{N}_t} \cdot \frac{R^{M}_t - R^{N}_t}{R^{FM}_t - R^{FN}_t}$$

$$\Delta_t \cdot \frac{R^{M}_t - R^{N}_t}{R^{FM}_t - R^{FN}_t} = \frac{\bar {R}_t - (R^{FN}_t + R^{{\pi}N}_t)}{R^{FM}_t - R^{FN}_t}$$

$$\Delta_t \cdot \frac{R^{M}_t - R^{N}_t}{R^{FM}_t - R^{FN}_t} = \nabla_t - \frac{R^{{\pi}N}}{R^{FM}_t - R^{FN}_t}$$

\begin{equation} \label{eq:decomposition}
    \Delta_t = (\nabla_t - \text{IE}) \cdot \xi_t
\end{equation}

where $\Delta_t$ corresponds to the theoretical $\Delta$ metric, $\nabla_t$ represents the gains in $\Delta$ derived from agent cooperation, IE captures the influence of inflation on economic rewards, and $\xi_t$ can be interpreted as the ratio of theoretical growth and inflation over the Nash and Monopoly equilibria.

Figure \ref{fig:decomposition} presents the breakdown of the $\Delta_t$ metric according to equation \ref{eq:decomposition}:

\begin{figure}[h]
  \centering
  \includegraphics[width=12cm]{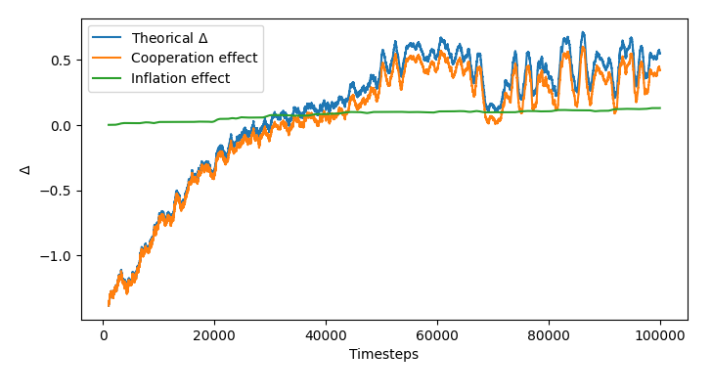}
  \caption{$\Delta_t$ decomposition into price and cooperation effects.}\label{fig:decomposition}
\end{figure}

\section{Sensitivy Analysis Results} \label{appendix:sensitivity}

To further investigate the impact of inflation on market competitiveness, we conduct a sensitivity analysis by varying key model parameters. Results are reported in Tables~\ref{tab:sensitivity_insample} and~\ref{tab:sensitivity_outsample}, which display outcomes for the in-sample and out-of-sample configurations, respectively. Table~\ref{tab:sensitivity_comparison} reports the results of an independent two-sample \textit{t}-test comparing both samples. The analysis reveals that configurations involving a larger number of agents are associated with increased market competitiveness, as evidenced by a significant decline in the coefficient~$\mu$.

\begin{table}[h!]
\centering
\caption{Sensitivy Results: In-sample environment}\label{tab:sensitivity_insample}
\begin{tabular}{|ccccc|}
\hline
Configuration & N & $\mu$ & $\sigma$ & Effect Size \\
\hline
$\gamma = 0.7$ & 50 & 0.3394 & 0.0023 & Very small \\
$\gamma = 0.8$ & 50 & 0.3392 & 0.0027 & Small \\
$\rho = 0.003$ & 50 & 0.3388 & 0.0059 & Medium \\
$\rho = 0.002$ & 50 & 0.3386 & 0.0040 & Medium \\
$k = 25$ & 50 & 0.3370 & 0.0170 & Very small \\
$lr = 0.02$ & 50 & 0.3202 & 0.0014 & Very small \\
Base experiment & 50 & 0.3156 & 0.0548 & - \\
$k = 10$ & 50 & 0.2964 & 0.1871 & Small \\
$lr = 0.03$ & 50 & 0.2940 & 0.0063 & Small \\
No Inflation & 50 & 0.2746 & 0.1664 & Small \\
$N = 5$ & 50 & 0.1300 & 0.0000 & Huge \\
$N = 3$ & 50 & 0.0506 & 0.0042 & Huge \\
\hline
\end{tabular}
\label{tab:entrenamiento}
\end{table}

\begin{table}[h!]
\centering
\caption{Sensitivy Results: Out-of-sample environment}\label{tab:sensitivity_outsample}
\begin{tabular}{|ccccc|}
\hline
Configuration & N & $\mu$ & $\sigma$ & Effect Size \\
\hline
$\gamma = 0.7$ & 50 & 0.3312 & 0.0607 & Very small \\
$\gamma = 0.8$ & 50 & 0.3396 & 0.0019 & Small \\
$\rho = 0.003$ & 50 & 0.3228 & 0.0851 & Very small \\
$\rho = 0.002$ & 50 & 0.3222 & 0.0880 & Very small \\
$k = 25$ & 50 & 0.3118 & 0.1012 & Very small \\
$lr = 0.02$ & 50 & 0.3296 & 0.0928 & Very small \\
Base experiment & 50 & 0.3258 & 0.0639 & - \\
$k = 10$ & 50 & 0.0818 & 1.0306 & Small \\
$lr = 0.03$ & 50 & 0.2670 & 0.1712 & Medium \\
No Inflation & 50 & 0.2920 & 0.1405 & Small \\
$N = 5$ & 50 & 0.1422 & 0.0303 & Huge \\
$N = 3$ & 50 & 0.0520 & 0.0127 & Huige \\
\hline
\end{tabular}
\label{tab:prueba}
\end{table}

\begin{table}[h!]
\centering
\caption{Sensitivy Results: Environment comparison}\label{tab:sensitivity_comparison}
\begin{tabular}{|cc|}
\hline
Configuration & $p(\mu_I = \mu_O)$ \\
\hline
$\gamma = 0.7$ & 0.3453 \\
$\gamma = 0.8$ & 0.4050 \\
$\rho = 0.003$ & 0.1909 \\
$\rho = 0.002$ & 0.1946 \\
$k = 25$ & 0.0805 \\
$lr = 0.02$ & 0.4774 \\
Base Experiment & 0.3937 \\
$k = 10$ & 0.1534 \\
$lr = 0.03$ & 0.2707 \\
No Inflation & 0.5735 \\
$N = 5$ & 0.0066 \\
$N = 3$ & 0.4650 \\
\hline
\end{tabular}
\label{tab:p_valores}
\end{table}

\end{document}